\documentclass[11pt]{article}

\usepackage[preprint]{acl}

\usepackage{times}
\usepackage{latexsym}

\usepackage[T1]{fontenc}

\usepackage[utf8]{inputenc}

\usepackage{tabularx}

\usepackage{microtype}

\usepackage{inconsolata}

\usepackage{graphicx}

\title{On the limited utility of parallel data for learning shared multilingual
representations}

\author{Julius Leino \\
  University of Helsinki \\
  \texttt{leino.julius2@gmail.com} \\\And
  Jörg Tiedemann \\
  University of Helsinki \\
  \texttt{jorg.tiedemann@helsinki.fi} \\}

\begin{document}
\maketitle
\begin{abstract}
Shared multilingual representations are essential for cross-lingual tasks and knowledge transfer across languages. This study looks at the impact of parallel data, i.e. translated sentences, in pretraining as a signal to trigger representations that are aligned across languages. We train reference models with different proportions of parallel data and show that parallel data seem to have only a minimal effect on the cross-lingual alignment. Based on multiple evaluation methods, we find that the effect is limited to potentially accelerating the representation sharing in the early phases of pretraining, and to decreasing the amount of language-specific neurons in the model. Cross-lingual alignment seems to emerge on similar levels
even without the explicit signal from parallel data.\footnote{Code for the experiments is available at \url{https://github.com/shiftleino/limited-utility-of-parallel-data}.}
\end{abstract}

\section{Introduction}
A desirable property of multilingual language models is the cross-lingual sharing of representations, which enables cross-lingual transfer \cite{Pires2019:mBERT, Artetxe2019:FrozenEnglishTransfer, conneau2020:emerging}, a phenomenon where representations learned for one language can be leveraged to boost performance in another. A common method thought to improve this sharing is to add parallel data, that is, translated texts, typically aligned at the sentence level, to the pretraining corpus of multilingual language models \cite{Conneau2019:TLM, Anil2023:PaLM2, Luukkonen2024:Poro, gemmateam2025:gemma3}. However, while intuitively attractive, it is still unclear to what extent parallel data actually helps in representation sharing.

In this paper, we systematically assess the assumption that including parallel data in the pretraining corpus increases the level of representation sharing across languages. To this end, we train from scratch four 1.4 billion-parameter transformer language models on a 200 billion token multilingual corpus with varying amounts of parallel data: 0\%, 1\%, 2\%, and 5\%. We then assess the level of representation sharing in the models using principal component projections, cosine similarity, projection weighted canonical correlation analysis (PWCCA), language-specific neuron analysis, and cross-lingual control vectors.

\begin{figure*}
    \includegraphics[width=0.9\textwidth]
    {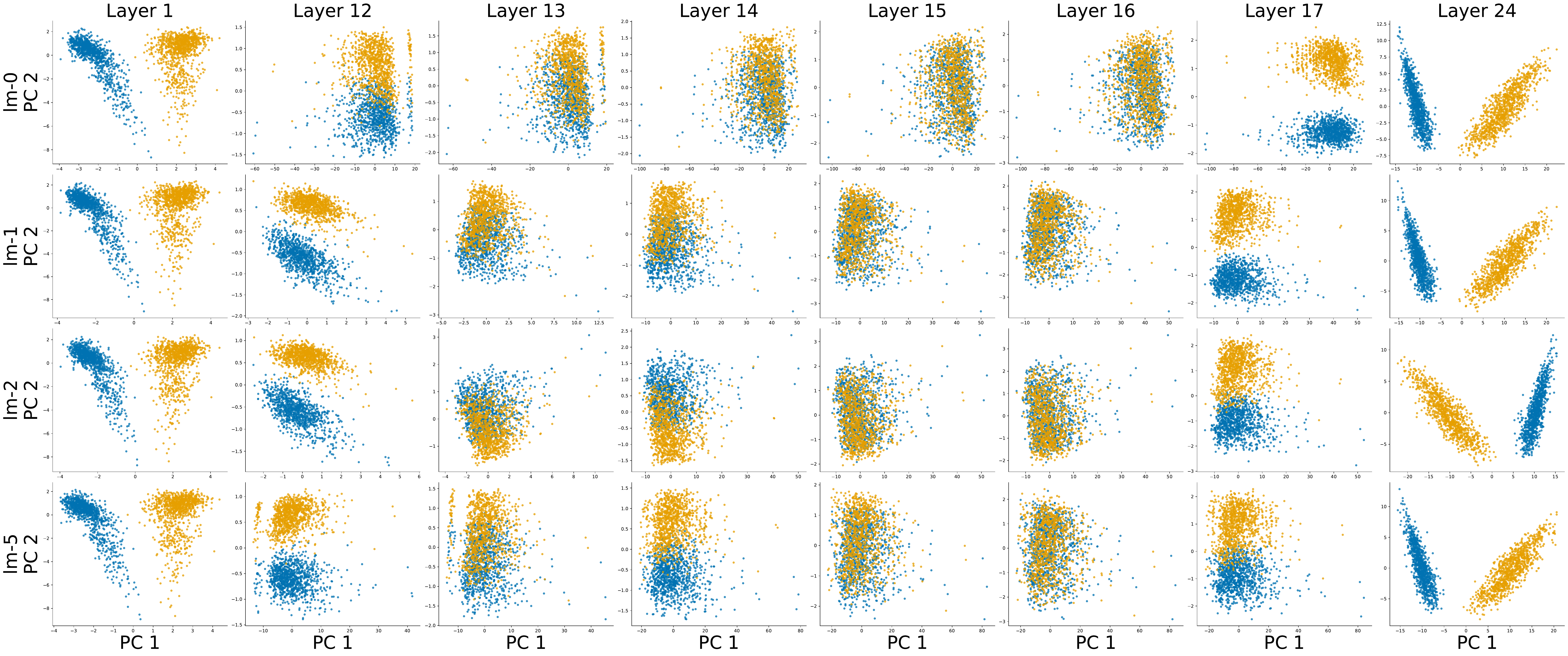}
    \centering
  \caption{English and Finnish translation sentence pair representations of each model projected to the first two principal components. Finnish sentences are in blue and English sentences in yellow. Each model is exposed to different amounts of parallel data from no parallel data in lm-0 (top row) to 5\% parallel data in lm-5 (bottom row).}
  \label{fig:projections-fully}
\end{figure*}

The primary findings of this study are as follows:
\begin{itemize}
    \item Our results across the different methods show that a partially shared representation space exists in all of the trained models, including the model with no parallel data added to the training corpus. The sharing of representations is strongest in the middle layers of all the models (Figure~\ref{fig:projections-fully}), aligning well with the observations of previous research on the middle layers containing more semantic representations \cite{Wendler2024:LLamasEnglish, kojima-etal-2024-multilingual}.
    \item We discover that the amount of parallel data in the pretraining corpus has little to no effect on the overall level of cross-lingual alignment in the fully trained models, with most of the evaluation methods showing no significant relationship with the amount of parallel data used in training. Instead, the effects of parallel data are limited to potentially accelerating the representation sharing during the early phase of the pretraining and to decreasing the number of language-specific neurons in the models.
\end{itemize}

\section{Background}
Prior research has suggested that multilingual language models develop shared cross-lingual representations. Numerous studies have, for example, shown that multilingual language models are capable of cross-lingual transfer, where a pretrained model trained for a task in one language achieves performance improvements in the same task in another language \cite{Pires2019:mBERT, Artetxe2019:FrozenEnglishTransfer, conneau2020:emerging}. Similar results have also been obtained from more direct analysis of the representations. For example, a study found that Llama~2 uses English as a pivot language when translating sentences from one language to another \cite{Wendler2024:LLamasEnglish}, while another study found that for many multilingual language models, neurons activating with a single language were predominantly located in the initial and final layers \cite{kojima-etal-2024-multilingual}. Furthermore, research has shown that the model output language can easily be changed using simple control vectors \cite{leino2025:xlingual-ctrl-vectors}, suggesting a disentangled language direction from the other concept directions. 

Beyond merely exploring how multilingual language models internally represent various languages, prior research has also examined methods to align these representations, often involving the usage of parallel data, i.e., aligned translated text, typically in the form of sentence pairs. In pretraining, parallel data has been involved, for example, with some auxiliary training objectives, such as a machine translation objective \cite{Kale2021:nmT5}. A common technique has also been to simply add a small portion of concatenated parallel sentences to the pretraining mix of language models without any additional training objective \cite{Conneau2019:TLM, Anil2023:PaLM2, Luukkonen2024:Poro, gemmateam2025:gemma3}. Intuitively, the attention blocks in the transformer architecture could provide the necessary communication mechanism between the translation sentences for the cross-lingual signal, thus improving cross-lingual representation sharing in the models. However, it is not yet fully understood whether this actually occurs in practice and how dependent it is on the amount of parallel data used during pretraining.

\section{Model training}
To study the effects of parallel data, we train four language models of 1.4 billion parameters using a bilingual corpus of 200 billion tokens with varying proportions of parallel data, using the training setting described in the following sections.

\subsection{Language models}
For the models, we closely follow the open source OLMo decoder-only transformer language model architecture \cite{groeneveld2024:olmo}. Our version of the model has 24 transformer layers in total, a residual stream of \(2048\) dimensions, and 16 attention heads with dimension \(128\) (see Table \ref{tab:model-hyperparams} in the Appendix~\ref{sec:appendix3} for full details).

\subsection{Training data}
To train each model, we construct a 200 billion token training corpus consisting of tokens in two languages, English and Finnish. Due to English and Finnish forming a typologically distant language pair, this setting provides a robust foundation for the generalization of the results to more similar language pairs. For each of the four models, we dedicate a different proportion of the data to translated sentence pairs (name of the model in parenthesis): 0\% (\textbf{lm-0}), 1\% (\textbf{lm-1}), 2\% (\textbf{lm-2}), and 5\% (\textbf{lm-5}).

We used a skewed language mix of 80\% tokens in English and 20\% in Finnish, with the assumption of parallel data having an equal split between the languages and then adjusting the monolingual corpus sizes to preserve this ratio. The distributions of the training corpora for the four models are shown in Table~\ref{tab:lang-allocation}.

\begin{table}[t]
  \begin{center}

  \begin{tabular}{l|rrr}
    \hline
    \textbf{Model} & \textbf{English} & \textbf{Finnish} & \textbf{Parallel} \\
    \hline
    lm-0 & \(160 \times 10^9\) & \(40 \times 10^9\) & 0 \\
    lm-1 & \(159 \times 10^9\) & \(39 \times 10^9\) & \(2 \times 10^9\) \\
    lm-2 & \(158 \times 10^9\) & \(38 \times 10^9\) & \(4 \times 10^9\) \\
    lm-5 & \(155 \times 10^9\) & \(35 \times 10^9\) & \(10 \times 10^9\) \\\hline
  \end{tabular}
  \caption{The distributions of the training data in tokens for the four language models.}
  \label{tab:lang-allocation}
  \end{center}
\end{table}

We utilized the FineWeb \cite{penedo2024:fineweb} and FineWeb2 \cite{penedo2025:fineweb2} datasets as the sources for our English and Finnish monolingual corpora, respectively, subject to additional custom quality filtering and upsampling steps detailed in Appendix~\ref{sec:appendix5}. For parallel data, we utilized three separate datasets from the OPUS collection \cite{tiedemann2012:opus}: HPLT\footnote{\url{https://hplt-project.org/}} (29,067,875 Finnish-English sentence pairs) \cite{de-gibert2024:hplt}, CCMatrix (35,982,562 sentence pairs) \cite{ Fan2020:MetaParallel, schwenk2021:ccmatrix}, and the 2024 version of the OpenSubtitles dataset\footnote{\url{http://www.opensubtitles.org/}} (110,093 sentence pairs) \cite{tiedemann2016:obensubtitles}. Our final parallel dataset contains a total of 65,160,530 parallel sentences, which results in approximately 4.5B tokens. We then up-sample this dataset to reach the required amount. To control for potential domain differences, a fixed baseline of 5\% of the full training corpora is sourced from the parallel dataset for all the models.

Instead of feeding the parallel data only in a single format, we apply to each parallel pair a format sampled from 60 different instruction or completion formats covering both English and Finnish (example in Appendix~\ref{sec:appendix1}). The intuition is that the variability in the format helps the model avoid overfitting to the specific format structure and instead allows learning solely the relationships between the languages.

To tokenize the collected text corpus, we train a SentencePiece tokenizer \cite{kudo2018:sentencepiece} using Byte Pair Encoding (BPE) \cite{sennrich2016:bpe} with a balanced corpus and a vocabulary size of 50,280 tokens.

\subsection{Training setup}
We trained the models largely following the codebase used for training the OLMo models \cite{groeneveld2024:olmo}, utilizing the LUMI supercomputer infrastructure\footnote{\url{https://lumi-supercomputer.eu}} and a context window of 2048 tokens. To avoid any unintended cross-lingual signals during the training, we did not concatenate sequences across the three corpora into same chunks, but still shuffled the chunks in the batches. A detailed description of the training configuration is outlined in Appendix~\ref{sec:appendix6}.

\section{Evaluation experiments}
We use five methods in total to evaluate the level of representation sharing in the models to obtain a comprehensive view on the potential effects.

\subsection{Principal component projections}
Our first method involves visualizing the representation space in two dimensions, similar to many other studies on multilingual language models \cite{marks2024:GeometryOfTruth, rimsky2024:CAA}. More specifically, we use the dev-split of the Finnish-English subset of the FLORES+ machine translation dataset \cite{nllb2024:flores}, which contains 997 translated sentences in total, run these sentences through the model and collect the mean pooled representations of each sentence from the residual stream after each layer of the network. After obtaining the representations for each sentence, we stack the representations for both languages and apply principal component analysis (PCA) to the resulting dataset. We then project the sentence representations to the first two principal components and visualize them for all layers. If the Finnish and English sequences align visually, the primary sources of variation are not language-specific, indicating that the sequences share some common representations. This will, therefore, provide first evidence of the existence of shared representations.

\subsection{Cosine similarity of parallel sentences}
To directly compare the representations across languages, we compute the average cosine similarities between representations of 2000 translation sentences sampled from the OPUS WMT-News dataset \cite{tiedemann2012:opus}. Similarly as with the projections, we use the mean pooled representations extracted from the residual stream after each layer of the model. Since cosine similarity measures the angular closeness of vectors, we expect higher scores between the parallel pairs when the representation space of the model is more cross-lingually aligned. We also compute the average cosine similarities between random sentences across the parallel data to act as a reference point. More specifically, we compute for each representation of Finnish sentence in the dataset the cosine similarity with 50 randomly sampled English sentences, take the average over these 50 cosine similarities, and finally take the average over all the Finnish sentences, thus resulting in one score representing the baseline cosine similarity.

\subsection{Projection weighted canonical correlation analysis}
We also perform projection weighted canonical correlation analysis (PWCCA) \cite{Morcos2018:PWCCA} between 20,000 translation pairs, similar to a study analyzing multilingual representations of mBERT \cite{Singh2019:PWCCA}. We again sample sentence pairs from the OPUS WMT-News dataset \cite{tiedemann2012:opus}. The advantage of PWCCA is that it is invariant to linear transformations of the representations, thus making it suitable for analyzing multilingual representations, where some language-specific linear transformations might be applied on top of the shared representation space.  Intuitively, in case the parallel data helps in cross-lingual representation sharing, we would expect higher PWCCA scores with larger amounts of parallel data used in training. To calculate the PWCCA scores, we use the implementation of the original paper~\cite{Morcos2018:PWCCA} available on GitHub\footnote{\url{https://github.com/google/svcca/blob/master/pwcca.py}}.

\subsection{Distribution of language-specific neurons}
To assess the representation space from the point of view of neuron activations, we identify language-specific neurons and assess their amount and distribution across layers using a method proposed by \citet{kojima-etal-2024-multilingual}. More specifically, we will first collect the neuron activation values for each neuron in the model when feeding the model with both English and Finnish text from the Finnish-English development subset of the FLORES+ dataset \cite{nllb2024:flores}. We then take the mean activation value of each neuron over the tokens of each sequence, capturing one activation value for each neuron per sequence in the dataset. Using these activation values and language labels per sentence, we regard the activation values of each neuron as the predicted scores for the language of an input sequence, and the original language of the sequence as the ground truth. We then measure the average precision (AP), which should capture the language sensitivity of the neuron. Having obtained the AP score for each neuron, we use these scores as a measure of language specificity in two analyses: 1) we plot the distribution of the top k scoring neurons across layers to understand the overall distribution of language-specific neurons and 2) we bucket neurons by AP thresholds for better model-to-model comparability.
We expect the number of less language-specific neurons to be higher, i.e. more neurons to be cross-lingually shared, when more parallel data is used in training.

\subsection{Effectiveness of language control vectors}
We also evaluate the effectiveness of language control vectors \cite{leino2025:xlingual-ctrl-vectors} that we employ to force output in Finnish for the completions of 200 English context sentences sampled from the ROCStories dataset \cite{Mostafazadeh2016:ROCStories}. 
When employing the control vectors, we use a similar approach to that of \citet{leino2025:xlingual-ctrl-vectors}, creating the control vectors from representations of 997 Finnish-English parallel sentences sampled from FLORES+ \cite{nllb2024:flores}, conducting a hyperparameter search for the best scaling factor \textit{a}, and applying these vectors to the residual stream after each layer of the network at each token position of both the prompt and generated tokens.

For each story in the dataset, we let the models generate five continuations and evaluate using an LLM-judge, Mistral Medium 3 \cite{mistral2025:Medium3}, both the Finnish fluency and the coherence of the completion. More specifically, the task of the LLM-judge is for both of the metrics to map each completion to one of four categories provided in the prompt, with the first category representing a complete failure, while the fourth category a perfectly valid completion with the rest of the categories falling in between (Appendix~\ref{sec:appendix2}). Furthermore, to provide baselines for both of these metrics, we also generate completions without control vectors for the same English stories and for the corresponding Finnish translations obtained from Gemini 2.5 Pro \cite{GeminiTeam2025:Gemini2.5}. Intuitively, we expect to get better fluency and coherence scores with more aligned representations because the language direction captured by the control vectors becomes more decoupled from the other representations.

\section{Results}
\subsection{Fully trained models}
The visualizations of the projections to the first two principal components show that even the model without parallel data achieves some degree of cross-lingual alignment in the middle layers, with the English and Finnish data points heavily overlapping each other, as shown in Figure~\ref{fig:projections-fully}. The overlap seems to happen in the same middle layers across all the models, with only some variation in the degree of overlap in the layers. Overall, these results provide the first evidence that cross-lingual representation sharing is occurring in each of the models, regardless of parallel data.

\begin{figure}[t]
  \includegraphics[width=\columnwidth]{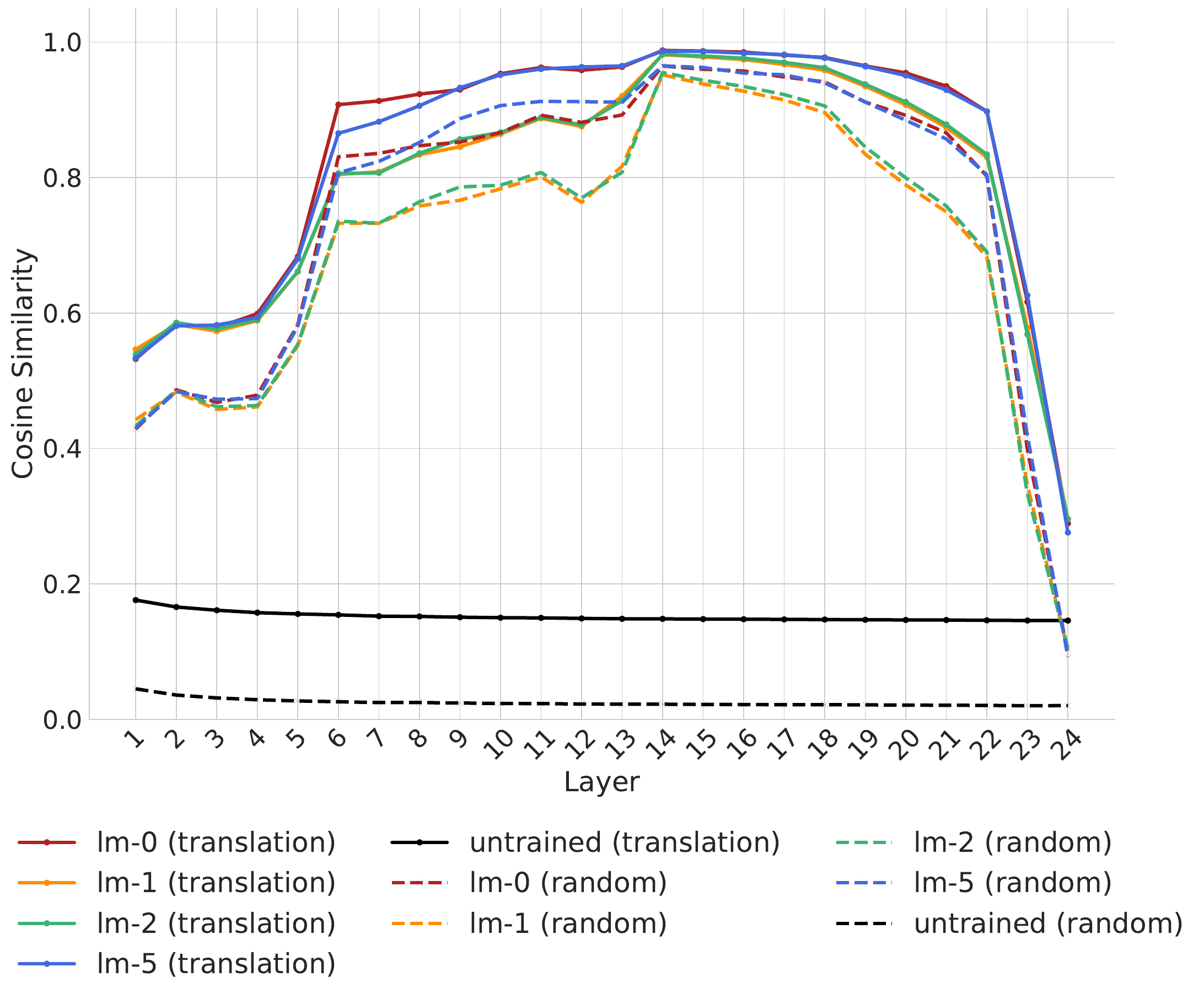}
  \caption{The average cosine similarities across the layers for the fully trained models and the untrained model. \textbf{Translation} refers to the average cosine similarity between the Finnish-English translation pairs, whereas \textbf{random} refers to the average cosine similarity between random Finnish and English sentence pairs.}
  \label{fig:cosine-similarity-abs}
\end{figure}

From Figure~\ref{fig:cosine-similarity-abs}, we can see that for each model the average cosine similarity of translation pairs increases quickly between layers 4 and 6, stays high during the middle layers, and decreases sharply after layer 22. Although these results alone indicate that the translation pairs share common representations in the middle layers, we can see that the cosine similarities of Finnish sentences with random English sentences also follow this pattern. Therefore, instead of simply sharing semantic representations between languages and therefore pushing the cosine similarity of translation pairs close to \(1.0\), the model seems to push \textit{all} representations into a small region in the representation space. This phenomenon is a well-known property of many language models, often referred to as anisotropy \cite{ethayarajh2019:anisotropy}.

Furthermore, while we can still see from Figure~\ref{fig:cosine-similarity-abs} that the average cosine similarity between translation pairs seems to be significantly higher for all models compared to random sentence pairs, even the untrained model seems to show a clear gap between the average cosine similarities of translation pairs and random Finnish-English sentence pairs. Therefore, it remains difficult to obtain a definite answer as to whether the higher cosine similarity of translation pairs is due to representation sharing or a residual effect from token embeddings that might be shared across languages through, e.g., named entities. 

\begin{figure}[t]
  \includegraphics[width=\columnwidth]{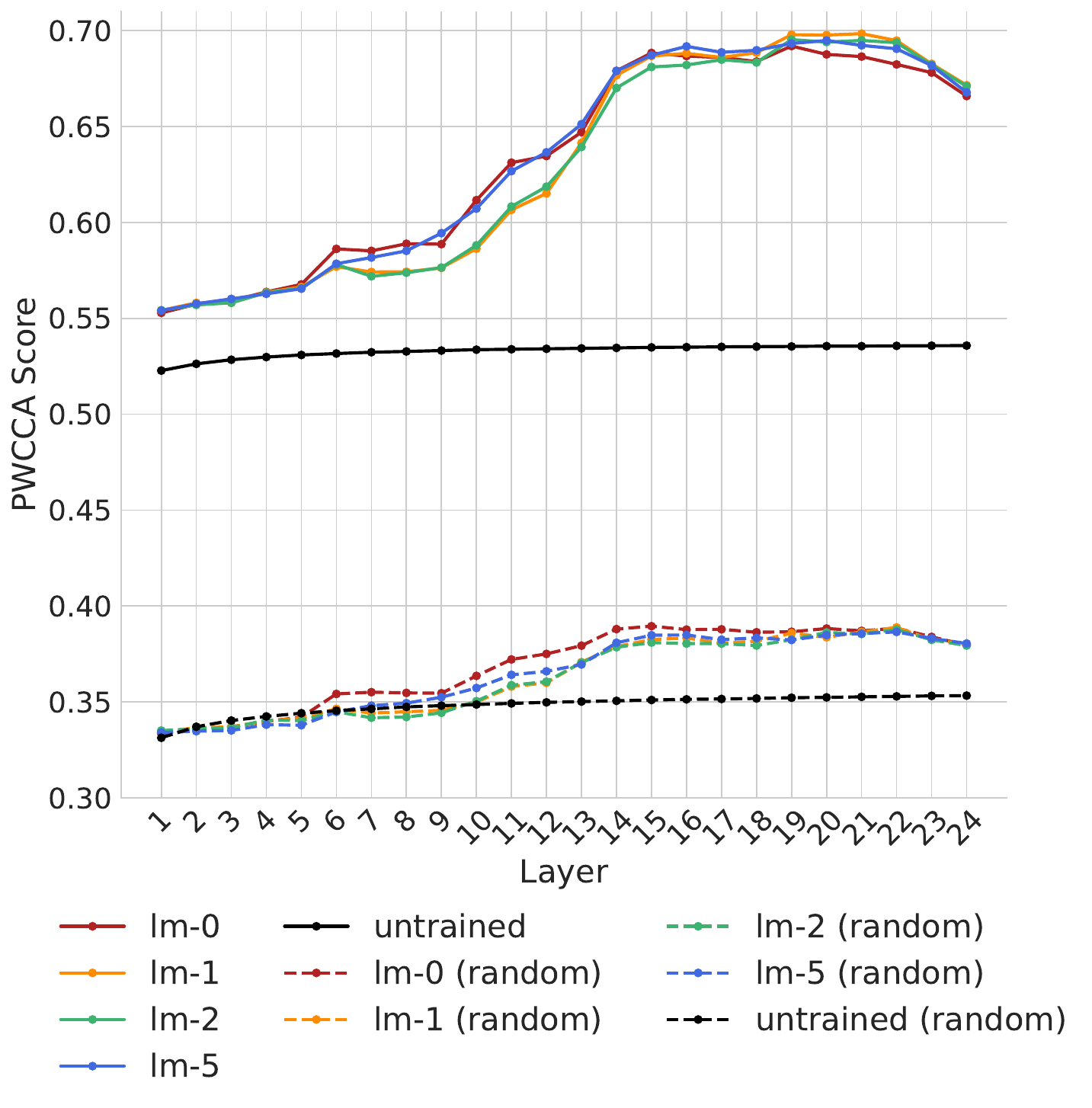}
  \caption{The PWCCA scores for the final checkpoints of all the models and the untrained model. \textbf{Random} refers to the scores obtained when shuffling the parallel dataset, thus breaking the possible correlation.}
  \label{fig:pwcca-fully}
\end{figure}

However, as shown in Figure~\ref{fig:pwcca-fully}, the PWCCA scores are well beyond the scores of the untrained model and the scores with shuffled pairs, thus supporting the findings of the PCA projections that all models learn to share representations between Finnish and English. Furthermore, the PWCCA scores show a clear trend of increasing first slowly during the early layers, then rapidly in the early middle layers, and staying close to the peak value of 0.7 until layer 22, from where the scores then drop only slightly. Intuitively, these results indicate that in the early middle layers the different components of the models start to write increasing amounts of cross-lingually shared representations into the residual stream, thus making the linear transformations of the representations of the parallel pairs correlate more with each other.

Furthermore, the results do not show any relationship between the amount of parallel data in the training corpus and the PWCCA scores since each model obtains a similar peak score with a similar increasing pattern in the early middle layers. These results indicate that with our models, after around 200B tokens, the general level of alignment between languages is invariant to the amount of parallel data used in the pretraining.

\begin{figure}[t]
  \includegraphics[width=\linewidth]{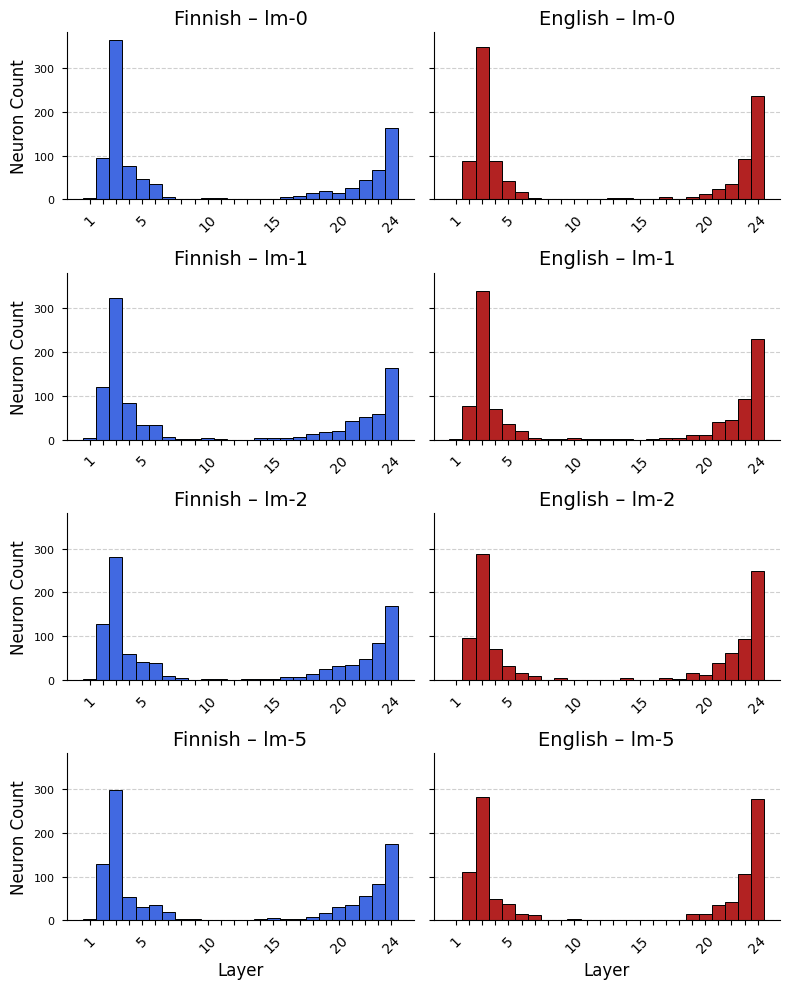}
  \caption{The distribution of the 1000 most language-specific neurons across the layers of each model.}
  \label{fig:neurons-all-fully}
\end{figure}

Figure~\ref{fig:neurons-all-fully} shows the distribution of the 1000 highest-scoring neurons for both Finnish and English across the models. From the figure, one can see that for all the models the most language-specific neurons are distributed mainly to the first few layers 
and the last few layers, with little to no language-specific neurons in the middle layers, aligning well with previous results on multilingual neurons \cite{kojima-etal-2024-multilingual}. The lack of language-specific neurons in the middle layers indicates that in these middle layers the models' neurons are more focused on shared features in the input representations rather than on features bound to either Finnish or English.

\begin{figure}[t]
  \includegraphics[width=\linewidth]{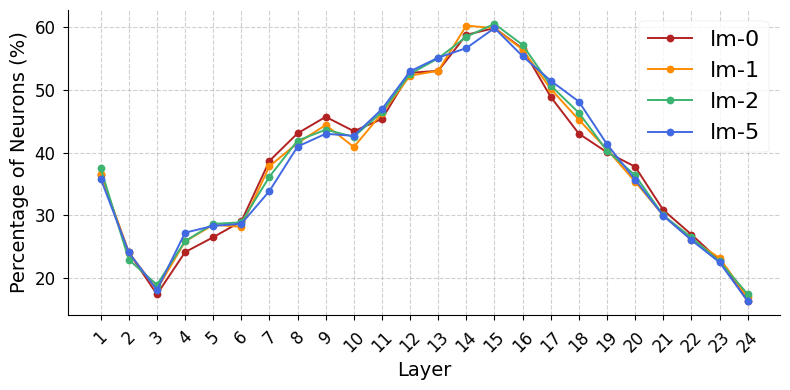}
  \caption{The percentage of neurons on each layer that score lower than or equal to 0.55 average precision in predicting both of the languages.}
  \label{fig:neurons-language-agnostic-fully}
\end{figure}

To better understand and compare the distributions of cross-lingually shared neurons, we visualize in Figure~\ref{fig:neurons-language-agnostic-fully} the percentage of neurons in each layer of the models that have an average precision of less than or equal to 0.55 for both Finnish and English. From the figure, we can see that for each model, the portion of these highly language-agnostic neurons peaks in layers 14 and 15 with close to 60\% of the neurons being language-agnostic. Furthermore, there seems to be no significant differences between the models in the proportions of language-agnostic neurons, indicating that increasing amounts of parallel data does not have a notable effect on the number of cross-lingually shared neurons.

Table~\ref{tab:neurons-total} shows the total sum of language-specific neurons with varying thresholds, where \textbf{lm-0} has the most language-specific neurons for both Finnish and English with the first three thresholds. Therefore, even though parallel data has limited effect on completely language-agnostic neurons, including such data seems to decrease the overall number of language-specific neurons, with the exception of most language-specific neurons.

\begin{table}
  \centering
  \begin{tabular}{lllll}
    \hline
    \textbf{Finnish} & \textbf{lm-0} & \textbf{lm-1} & \textbf{lm-2} & \textbf{lm-5} \\
    \hline
    >= 0.75 & \textbf{4666} & 4485 & 4538 & 4531\\
    >= 0.9 & \textbf{496} & 423 & 398 & 437\\
    >= 0.95 & \textbf{134} & 106 & 92 & 108\\
    >= 0.99 & 16 & \textbf{19} & 16 & 15\\
    \hline
    \textbf{English} & \textbf{lm-0} & \textbf{lm-1} & \textbf{lm-2} & \textbf{lm-5} \\
    \hline
    >= 0.75 & \textbf{3448} & 3208 & 3165 & 3201\\
    >= 0.9 & \textbf{317} & 274 & 243 & 279\\
    >= 0.95 & \textbf{92} & 78 & 67 & 78\\
    >= 0.99 & 8 & 10 & 9 & \textbf{11}\\
    \hline
    \textbf{Both} & \textbf{lm-0} & \textbf{lm-1} & \textbf{lm-2} & \textbf{lm-5} \\
    \hline
    <= 0.55 & 50,716 & 50,719 & \textbf{51,051} & 50,696\\
    \hline
  \end{tabular}
  \caption{\label{tab:neurons-total}
    The total number of neurons in each model with varying average precision thresholds. Finnish refers to average precision predicting Finnish, whereas English to average precision predicting English. The total number of neurons in each model is 132,096.
  }
\end{table}

Finally, Table~\ref{tab:language-control-fully} shows that the control vectors with the best scaling factor found during the initial hyperparameter search (\textit{a}=0.09) seem to be highly effective in changing the language of the completions from English to Finnish, obtaining mean fluency scores slightly below the maximum of 4 (grammatically perfect Finnish) and almost the same as when generating completions to the corresponding Finnish translations. However, there seems to be no positive correlation between fluency scores and the amount of parallel data. Furthermore, although controlled completions obtain lower coherence scores than the uncontrolled ones, the scores are still well above the baseline score of 1 (completely unrelated continuations to the preceding story context). Therefore, even when shifting the language to Finnish, all the models can generate completions that are somewhat related to the previous English context, suggesting cross-lingually shared representations (example shown in Table~\ref{tab:language-control-example} in Appendix~\ref{sec:appendix4}). Between the models, \textbf{lm-5} seems to obtain slightly higher coherence score than the rest of the models; however, as shown by the higher scores for the baseline continuations, this might just be due to better overall coherence of the model's outputs.

\begin{table}
  \centering
  \begin{tabular}{lcccc}
        \hline
         & \multicolumn{2}{c}{\textbf{Fluency}} & \multicolumn{2}{c}
         {\textbf{Coherence}}\\
         \hline
        \textbf{lm-0} & $\mu$ & $\sigma$ & $\mu$ & $\sigma$ \\
         \hline
     FI & 3.818 & 0.221 & 2.899 & 0.479\\
         EN & 1.008 & 0.052 & 3.123 & 0.425\\
         0.09 & 3.713 & 0.266 & 2.324 & 0.636\\
         \hline
         \textbf{lm-1} & $\mu$ & $\sigma$ & $\mu$ & $\sigma$ \\
         \hline
         FI & 3.827 & 0.211 & 2.92 & 0.478 \\
         EN & 1.002 & 0.020 & 3.147 & 0.448 \\
         0.09 & 3.705 & 0.270 & 2.231 & 0.663\\
         \hline
         \textbf{lm-2} & $\mu$ & $\sigma$ & $\mu$ & $\sigma$ \\
         \hline
         FI & 3.83 & 0.192 & 2.99 & 0.425 \\
         EN & 1.009 & 0.054 & 3.152 & 0.433\\
         0.09 & 3.694 & 0.273 & 2.343 & 0.623\\
         \hline
         \textbf{lm-5} & $\mu$ & $\sigma$ & $\mu$ & $\sigma$ \\
         \hline
         FI & 3.805 & 0.214 & 3.03 & 0.402\\
         EN & 1.005 & 0.042 & 3.175 & 0.426\\
         0.09 & 3.678 & 0.267 & 2.461 & 0.622\\
    \hline
  \end{tabular}
  \caption{\label{tab:language-control-fully}
  The mean ($\mu$) scores with standard deviations ($\sigma$) from the LLM-judge for fluency and coherence for the continuations to the 200 stories. 0.09 refers to the controlled continuations, FI to the Finnish baseline continuations without any control vectors applied, and EN to English baseline continuations.}
\end{table}

\subsection{Throughout training}
\begin{figure}[t]
  \includegraphics[width=\linewidth]{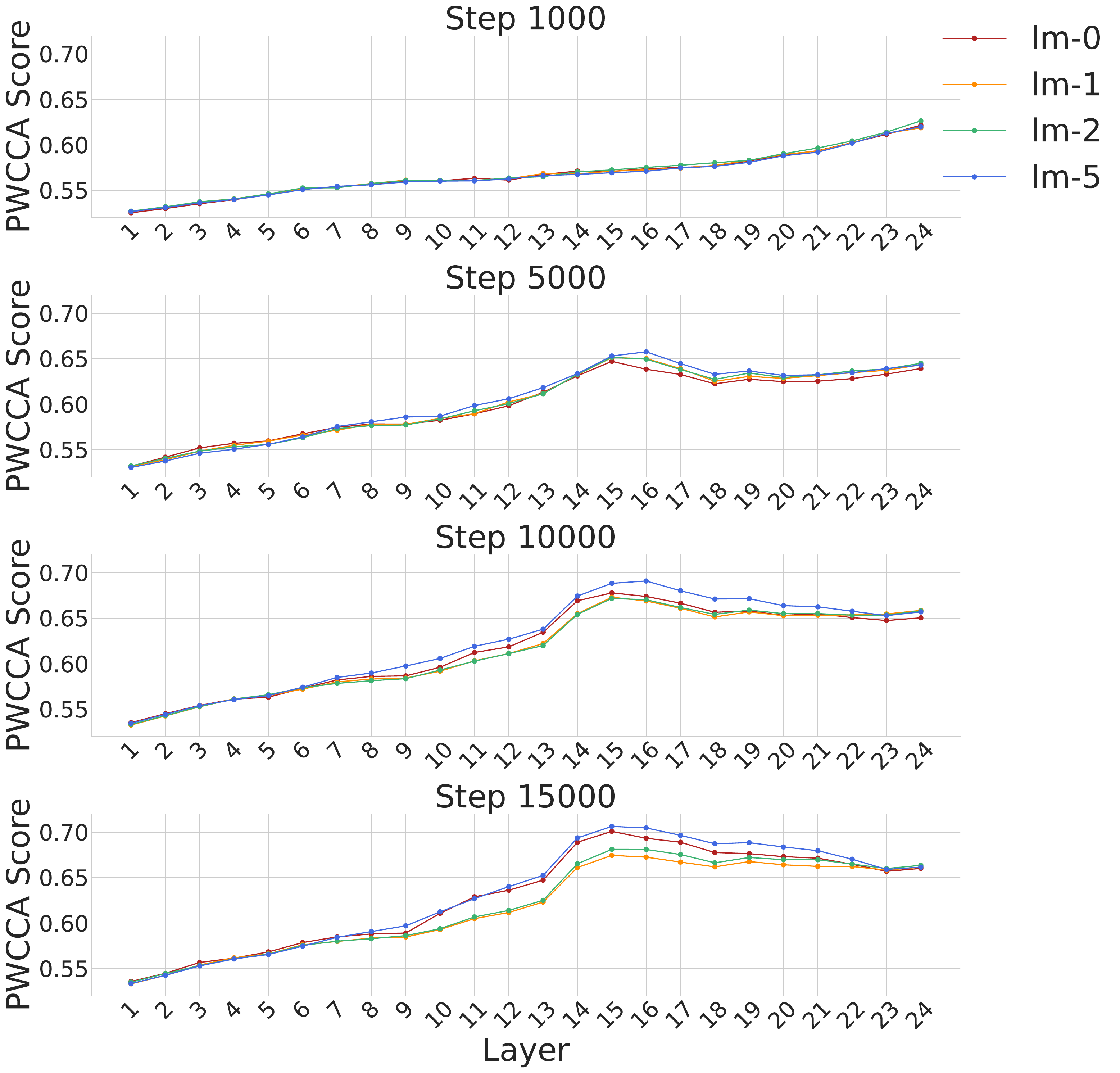}
  \caption{PWCCA scores computed across layers for early checkpoints of each model.}
  \label{fig:pwcca-early-checkpoints}
\end{figure}

We also evaluated the level of cross-lingual representation sharing in the checkpoints of the models  throughout the training. Based on the evaluation results, the cross-lingual alignment starts to emerge relatively early in the training. For example, Figure~\ref{fig:pwcca-early-checkpoints} shows that for all the models, the PWCCA scores start to increase very early in training with a bump in the scores appearing in the middle layers even as early as after 5000 steps. There seems to be no clear correlation with the amount of parallel data.

\begin{figure}[t]
  \includegraphics[width=\linewidth]{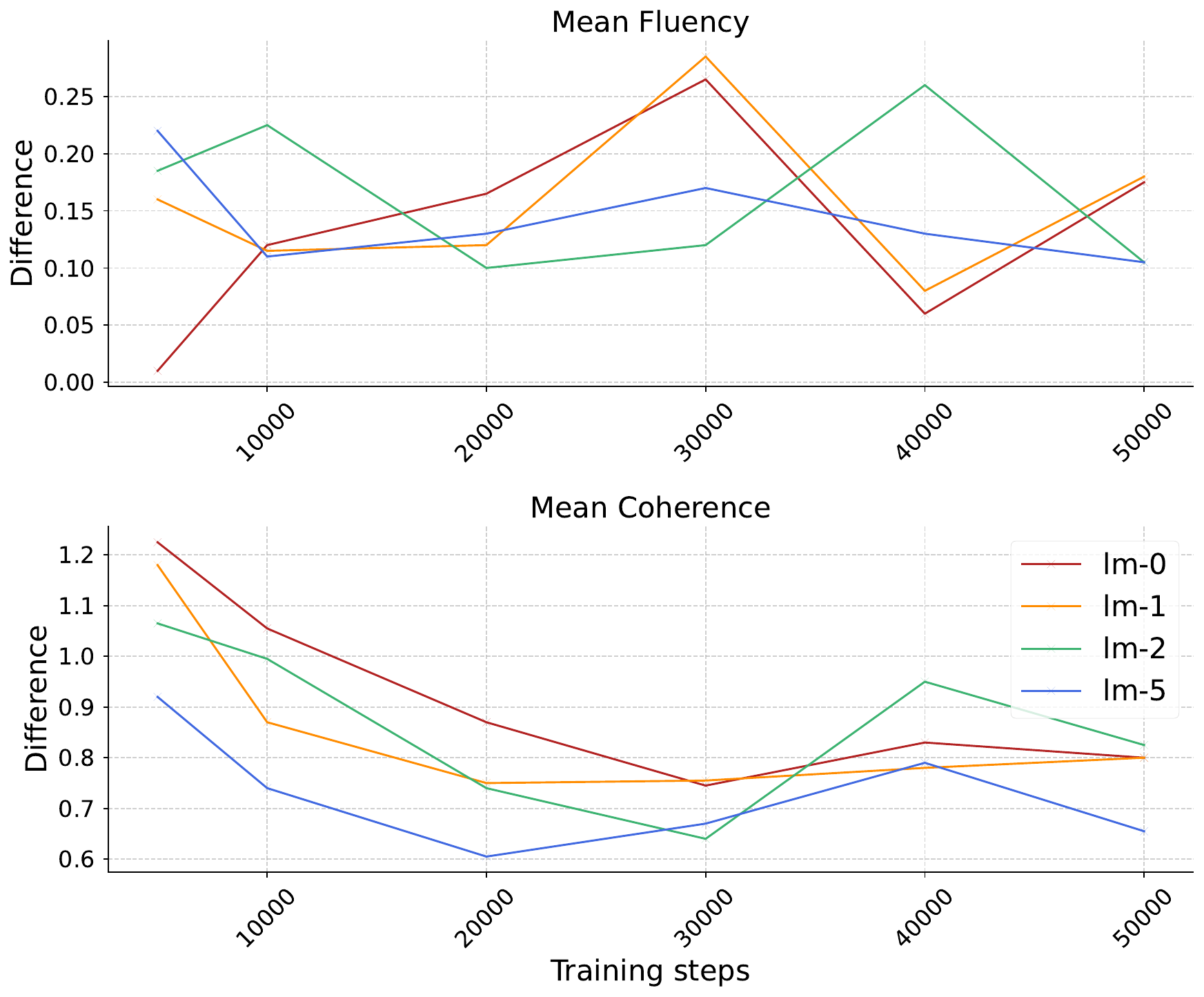}
  \caption{Difference in mean fluency and coherence scores between the controlled and Finnish baseline completions for checkpoints during training.}
  \label{fig:lang-control-early-checkpoints}
\end{figure}

On the other hand, the mean fluency and coherence scores in the language control experiment reveal a more noticeable distinction. To isolate the effects of general improvements in model capabilities, we normalized the scores by subtracting them from the Finnish baseline scores. Figure~\ref{fig:lang-control-early-checkpoints} shows these differences across the model checkpoints (\textit{a}=0.09). We can see that, while the differences in mean fluency scores do not show any trend, a clear distinction emerges in the coherence score differences. Assuming that perfectly aligned representations would produce a coherence difference of zero in the experiment, \textbf{lm-5} appears to have the most aligned representations during the initial 20,000 steps, while \textbf{lm-0} the least. These results suggest that adding parallel data helps in sharing representations in the early phases of the training.

\begin{figure}[t]
  \includegraphics[width=\linewidth]{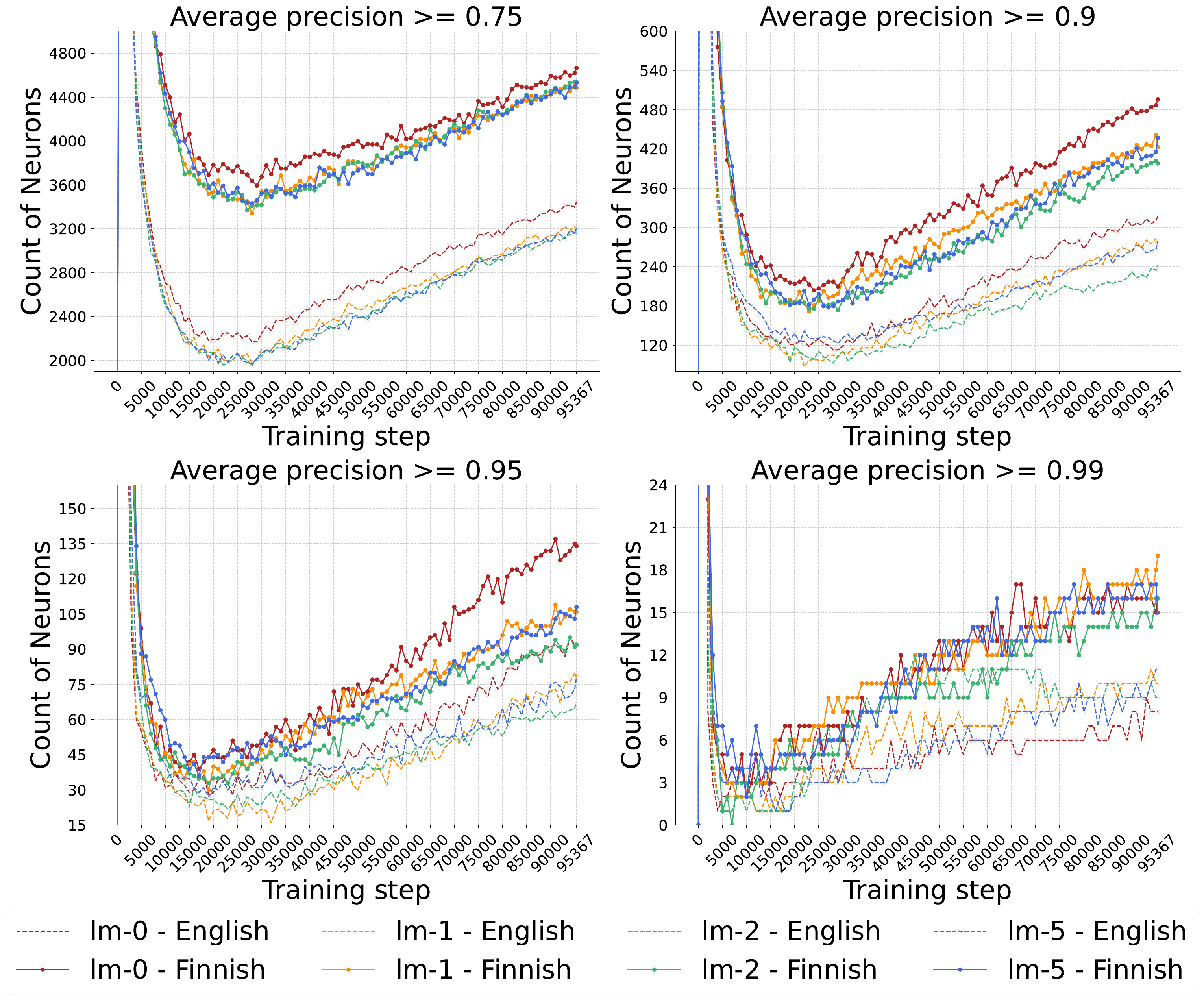}
  \caption{The development of the number of neurons given four thresholds for average precisions.}
  \label{fig:neurons-throughout-training}
\end{figure}

We also examined the number of language-specific neurons throughout the training with different average precision thresholds as shown in Figure~\ref{fig:neurons-throughout-training}. From the figure, we can see that \textbf{lm-0} has for most thresholds the most language-specific neurons throughout the training. However, what is even more interesting is that after an initial phase of rapid decrease, the number of language-specific neurons increases for all of the models throughout the whole training without showing any signs of convergence. This calls into question the use of the number of language-specific neurons as a measure for cross-lingual alignment, as the models seem to naturally converge toward a higher number of these neurons over time.

\section{Conclusions}
Overall, all of our experiments show evidence for the shared cross-lingual representation space emerging in the middle layers of all models, with the exception of cosine similarity, which provided somewhat unreliable results. This observation of cross-lingual representation sharing in the middle layers also aligns well with previous research \cite{Wendler2024:LLamasEnglish, kojima-etal-2024-multilingual}, and was therefore somewhat expected.

The more surprising result of the experiments is that, contrary to general belief, parallel data appears to have little effect on overall cross-lingual alignment in language models. In general, most of the evaluation experiments did not show any significant differences between the models. Therefore, it seems likely that some other factor encourages even the model without parallel data to share representations between languages, such as a limited parameter capacity budget \cite{conneau2020:emerging}. It is also possible that, despite the language filtering steps \cite{penedo2024:fineweb, penedo2025:fineweb2}, the monolingual corpora still contain parallel data similar to some other training corpora \cite{blevins-zettlemoyer-2022-language, Briakou2023:PalmBilingualism}. We leave the examination of these potentially overshadowing factors for future research.

Despite not finding evidence of parallel data helping in the representation sharing in fully trained models, we still observed some evidence of accelerated cross-lingual representation sharing during the early stages of pretraining. Furthermore, it seems that including parallel data to the training corpus might push some of the language-specific neurons to be less language-specific. Interestingly, we also found that the number of language-specific neurons increases across all models as training progresses.

To summarize, the results of this study show that while being an intuitively intriguing technique, including parallel data to the pretraining corpus does not seem to provide any major benefit to cross-lingual alignment of language models.

\section*{Limitations}
A primary limitation of this study is the restriction of our experiments to a single language pair, English and Finnish, which could limit the ability to generalize the findings to typologically closer pairs or other distant pairs. However, we selected English-Finnish specifically because it represents a "hard case" for cross-lingual alignment due to significant typological differences, thus making the results likely to generalize to other language pairs as well and the results not to be biased by the similarity of the languages.

Another limitation of the study is the scale of the trained language models. Constrained by computational resources, we trained only 1.4B-parameter models on 200B tokens, thus obtaining performance-wise limited models. Consequently, the generalizability of our findings to the large-scale models remains an open question as phenomena observed with small models might not always generalize to larger ones. However, training smaller models from scratch allowed us to exercise precise control over the data mixture, specifically the exact ratio of parallel tokens, thus allowing us to conduct the experiments in a more controlled manner.

Finally, our models ranged from leveraging 0\% to 5\% parallel data. While this range might be insufficient to observe potential effects that might emerge at higher ratios (e.g., 20\% or 50\%), we restricted our scope to this range to reflect realistic constraints in pretraining language models. For many language pairs, particularly low-resource ones, acquiring high-quality parallel data constituting more than 5\% of a massive pretraining corpus is often infeasible. Our objective was to evaluate the utility of parallel data within a regime of realistic scarcity, rather than in synthetic scenarios dominated by this parallel data.

\section*{Acknowledgments}
The research was supported by the Technology Industries of Finland Centennial Foundation.

\bibliography{custom}

\appendix

\section{Model architecture hyperparameters}
\label{sec:appendix3}
Table~\ref{tab:model-hyperparams} shows a detailed breakdown of the model architecture used in this study.

\begin{table}[h]
    \centering
    \begin{tabularx}{\columnwidth}{|l|X|}
        \hline
        \textbf{Hyperparameter} &  \textbf{Value} \\ \hline
        Parameters & \(1,420,296,192\) \\ \hline
        Non-embedding parameters & \(1,317,273,600\) \\ \hline
        Number of layers & \(24\) \\ \hline
        Model dimension & \(2048\) \\ \hline
        Attention type & Multi-headed attention \\ \hline
        Number of heads & \(16\) \\ \hline
        Q/K/V dimension & \(128\) \\ \hline
        Activation function & SwiGLU \\ \hline
        MLP dimension & \(5504\) \\ \hline
        Layer normalization & Non-parametric LayerNorm \\ \hline
        Position information & RoPE \\ \hline
        RoPE base \(\theta\) & \(10,000\) \\ \hline
        Vocabulary size & \(50,280\) \\ \hline
        Embedding size & \(50,304\) \\ \hline
    \end{tabularx}
    \caption{Hyperparameters for the model architecture used in this study.}
    \label{tab:model-hyperparams}
\end{table}

\section{Quality filtering and upsampling steps applied to the monolingual corpora}
\label{sec:appendix5}

As we noticed that the raw FineWeb2 dataset contained significant portions of low quality content, we applied our own keyword- and URL-based content filtering routines tailored for Finnish, resulting in the removal of approximately 8.8\% of the documents. To avoid a potential content imbalance between the Finnish and English corpora, we also then applied a similar content filtering step to the monolingual English corpus, leading to the removal of approximately 1.4\% of the documents. Due to the resulting Finnish dataset being only around 20B tokens, we further up-sampled the corpus accordingly to fit the Finnish data requirements of the models. This decision is supported by previous studies suggesting that training with the same data even up to four epochs has negligible effect on the downstream performance \cite{muennighoff2023:scalingdataconstrained}.

\section{Example instruction format}
\label{sec:appendix1}

\begin{quote}
Translate the English sentence '\{example\_en\}' into Finnish. Finnish: \{example\_fi\}.
\end{quote}

\section{Training hyperparameters and hardware setup}
\label{sec:appendix6}

For monolingual corpora, we concatenated the documents that were shorter than the context window of 2048 tokens together to fill the whole context window in one chunk, using special beginning of sequence \textbf{<s>} and end of sequence \textbf{</s>} tokens to separate the sequences. For parallel data, we concatenated the maximum number of formatted parallel sequences without overflowing the context window, again using the special tokens as separators, and applied padding to fill the remaining context window. To avoid any unintended cross-lingual signals, we kept all the datasets in separate chunks. However, we still applied shuffling to the chunks, thus allowing one batch to contain chunks from all the datasets. We trained the models with a batch size of approximately two million tokens (1024 chunks per batch), resulting in 95,367 total training steps.

For the optimizer, we used AdamW \cite{loshchilov2018decoupled} with beta coefficients of \(0.9\) and \(0.95\), and a weight decay coefficient of \(0.1\). For the learning rate, we employed a cosine learning rate scheduler with a warmup of 2000 steps and a maximum learning rate of \(2.0\times10^{-4}\) and a minimum learning rate of \(2.0\times10^{-5}\). We also employed gradient clipping to a maximum norm of \(1.0\).

From a hardware perspective, we trained the models with a cluster of 32 AMD Instinct MI250X GPUs (64 Graphics Compute Dies, GCDs) divided across 8 nodes (4 physical GPUs per node, 8 GCDs) hosted on the GPU partition of the LUMI supercomputer.

\section{LLM-judge evaluation categories}
\label{sec:appendix2}

\noindent \textbf{Fluency}
\begin{enumerate}
  \item The text is grammatically malformed to the point of being incomprehensible (gibberish), is not in Finnish, or contains only a few Finnish words in a sea of another language.
  \item The text is understandable and mostly in Finnish, but contains many significant grammatical errors or words from other languages. Grammatical errors may include fundamentally wrong verb conjugations or noun inflections, severe phrasing issues, or highly repetitive words or phrases.
  \item The text is fully in Finnish and grammatically correct for the most part, but contains minor, non-critical errors. Examples include unnatural phrasing that suggests a literal translation from another language (calque), occasional incorrect word inflections, or awkward word choices.
  \item The text is fully in Finnish and grammatically perfect, idiomatic, and reads as if written by a native Finnish speaker. All word choices, inflections, and structures are natural.
\end{enumerate}

\noindent \textbf{Coherence}
\begin{enumerate}
  \item The continuation is completely unrelated to the previous sentences with no semantically resembling words, concepts, or structure indicating of relatedness.
  \item The continuation is related to the overarching topic or themes of the previous sentences (e.g., 'dogs', 'shopping') but fails to connect to the specific events or characters of the sentences.
  \item The continuation correctly identifies and uses specific elements (characters, objects, locations) from the previous sentences. However, it contains some major logical inconsistencies, e.g., by mixing up the subject of some actions or presenting conflicting facts about the elements.
  \item The continuation provides a coherent continuation to the previous sentences without illogical or inconsistent elements.
\end{enumerate}

\section{Example text completions from language control experiment}
\label{sec:appendix4}

Table~\ref{tab:language-control-example} shows controlled Finnish text completions to an example English story context.

\begin{table}[t]
    \centering
    \begin{tabular}{c|p{0.75\columnwidth}}
        \hline
         \textbf{Input} & Kia had always had a distaste for mushrooms. But then one day she ordered pasta at a restaurant. The sauce was wonderful, and she asked what was in it. To her surprise, the waiter said it was mostly mushrooms!\\
         \hline
         \textbf{lm-0} & Kia ei ollut ainoa, joka oli ihastunut t\"ah\"an ruokaan. Pian muutkin asiakkaat alkoivat kertoa sy\"ov\"ans\"a sieniruokaa. Sitten Kia p\"a\"atti
         \textit{Kia wasn't the only one who was in love with this dish. Soon other customers started telling me they were eating mushroom food. Then Kia decided}\\
         \hline
         \textbf{lm-1} & Kia oli t\"aysin j\"arkyttynyt. H\"an ei ollut koskaan sy\"onyt mit\"a\"an niin hyv\"a\"a. Mik\"a ihme on t\"am\"a? Mutta siin\"a vaiheessa h\"an tajusi, ett\"a
         \textit{Kia was completely shocked. She had never eaten anything so good.
What the hell is this? But at that moment she realized that}\\
         \hline
         \textbf{lm-2} & Kia ei ollut t\"ast\"a moksiskaan, vaan ajatteli, ett\"a totta kai h\"an s\"oi tuon. Mutta sitten h\"an huomasi, ett\"a ruoka oli
         \textit{Kia wasn't at all surprised by this,
and thought that of course he ate that.
But then he noticed that the food
was}\\
         \hline
         \textbf{lm-5} & Kia ei ollut koskaan aiemmin ostanut ravintolasta mit\"a\"an muuta kuin ruokaa, joten h\"an ei ollut kovin yll\"attynyt kuullessaan, ett\"a ateria oli tehty sienist\"a
         \textit{Kia had never bought anything other than food from a restaurant before, so she wasn't too surprised to hear that the meal was made with mushrooms.}\\
         \hline
    \end{tabular}
    \caption{Example text completions from the language control vector experiment. English translation in italics.}
    \label{tab:language-control-example}
\end{table}

\end{document}